\def\year{2022}\relax
\documentclass[letterpaper]{article} 
\usepackage{aaai22}  
\usepackage{times}  
\usepackage{helvet}  
\usepackage{courier}  
\usepackage[hyphens]{url}  
\usepackage{graphicx} 
\usepackage{natbib}  
\usepackage{caption} 
\DeclareCaptionStyle{ruled}{labelfont=normalfont,labelsep=colon,strut=off} 
\usepackage{algorithm}
\usepackage{algorithmic}

\usepackage{newfloat}
\usepackage{listings}
\floatstyle{ruled}
\newfloat{listing}{tb}{lst}{}
\floatname{listing}{Listing}
\usepackage{amsmath}
\usepackage{amssymb}
\usepackage{gensymb}
\usepackage{mathtools}
\usepackage{bm}
\usepackage{multirow}
\usepackage{makecell}
\usepackage{adjustbox}
\usepackage{subcaption}

\newcommand{\unaryminus}{\scalebox{0.75}[1.0]{\( - \)}}
\newcommand{\shortequal}{\scalebox{0.75}[1.0]{\( = \)}\,}
\newcommand{\shortrarrow}{\scalebox{0.75}[1.0]{\( \rightarrow \)}\,}

\def\a{{\textbf{a}}} \def\b{{\textbf{b}}} \def\c{{\textbf{c}}}
\def\f{{\textbf{f}}} 
\def\m{{\textbf{m}}} \def\p{{\textbf{p}}} \def\s{{\textbf{s}}}
\def\t{{\textbf{t}}}  \def\x{{\textbf{x}}}

 \def\B{{\textbf{B}}} 
\def\P{{\textbf{P}}} 
\def\F{{\textbf{F}}} \def\M{{\textbf{M}}}

\def\one{{\textbf{1}}}
\def\bPhi{{\boldsymbol{\Phi}}} 
\def\bphi{{\boldsymbol{\phi}}} \def\bpsi{{\boldsymbol{\psi}}}

  \def\cI{{\mathcal{I}}}
\def\cL{{\mathcal{L}}} \def\cN{{\mathcal{N}}} 
\def\cT{{\mathcal{T}}}

\title{Texture Generation Using Dual-Domain Feature Flow\\ with Multi-View Hallucinations}
\author{
    Seunggyu Chang\textsuperscript{\rm 1}, Jungchan Cho\textsuperscript{\rm 2}, Songhwai Oh\textsuperscript{\rm 1}\\
}
\affiliations{
    \textsuperscript{\rm 1}Department of ECE, ASRI, Seoul National University\\
    \textsuperscript{\rm 2}School of Computing, Gachon University\\
    seunggyu.chang@rllab.snu.ac.kr, thinkai@gachon.ac.kr, songhwai@snu.ac.kr\\
    
}

\begin{document}

\maketitle

\begin{abstract}
We propose a dual-domain generative model to estimate a texture map from a single image for colorizing a 3D human model. When estimating a texture map, a single image is insufficient as it reveals only one facet of a 3D object. To provide sufficient information for estimating a complete texture map, the proposed model simultaneously generates multi-view hallucinations in the image domain and an estimated texture map in the texture domain. During the generating process, each domain generator exchanges features to the other by a flow-based local attention mechanism. In this manner, the proposed model can estimate a texture map utilizing abundant multi-view image features from which multi-view hallucinations are generated. As a result, the estimated texture map contains consistent colors and patterns over the entire region. Experiments show the superiority of our model for estimating a directly render-able texture map, which is applicable to 3D animation rendering. Furthermore, our model also improves an overall generation quality in the image domain for pose and viewpoint transfer tasks.
\end{abstract}

\section{Introduction}

Along with the increase of online activities recently, reconstructing a 3D avatar from photos becomes an important problem. To reconstruct a 3D avatar, we need a 3D mesh template for shape representation and a corresponding texture map for colorization. Traditionally, a 3D avatar reconstruction requires multiple image pairs consisting of diverse poses and viewpoints taken at a dedicated studio. However, trend has move on to reconstructing a 3D model from fewer or a single image.
Many works \citep{bogo2016keep,lassner2017unite,kanazawa2018end,pavlakos2018learning,varol2018bodynet,alldieck2019learning,natsume2019siclope,weng2019photo,gabeur2019moulding,saito2019pifu,kolotouros2019spin,choutas2020monocular,saito2020pifuhd} have been studied to resolve 3D human shape reconstruction from a single image, however, little has been studied for texture map reconstruction from a single image for a 3D model \citep{wang2019reidsupervised,lazova2019360}.

\begin{figure}[t]
    \centering
    \includegraphics[width=0.84\columnwidth]{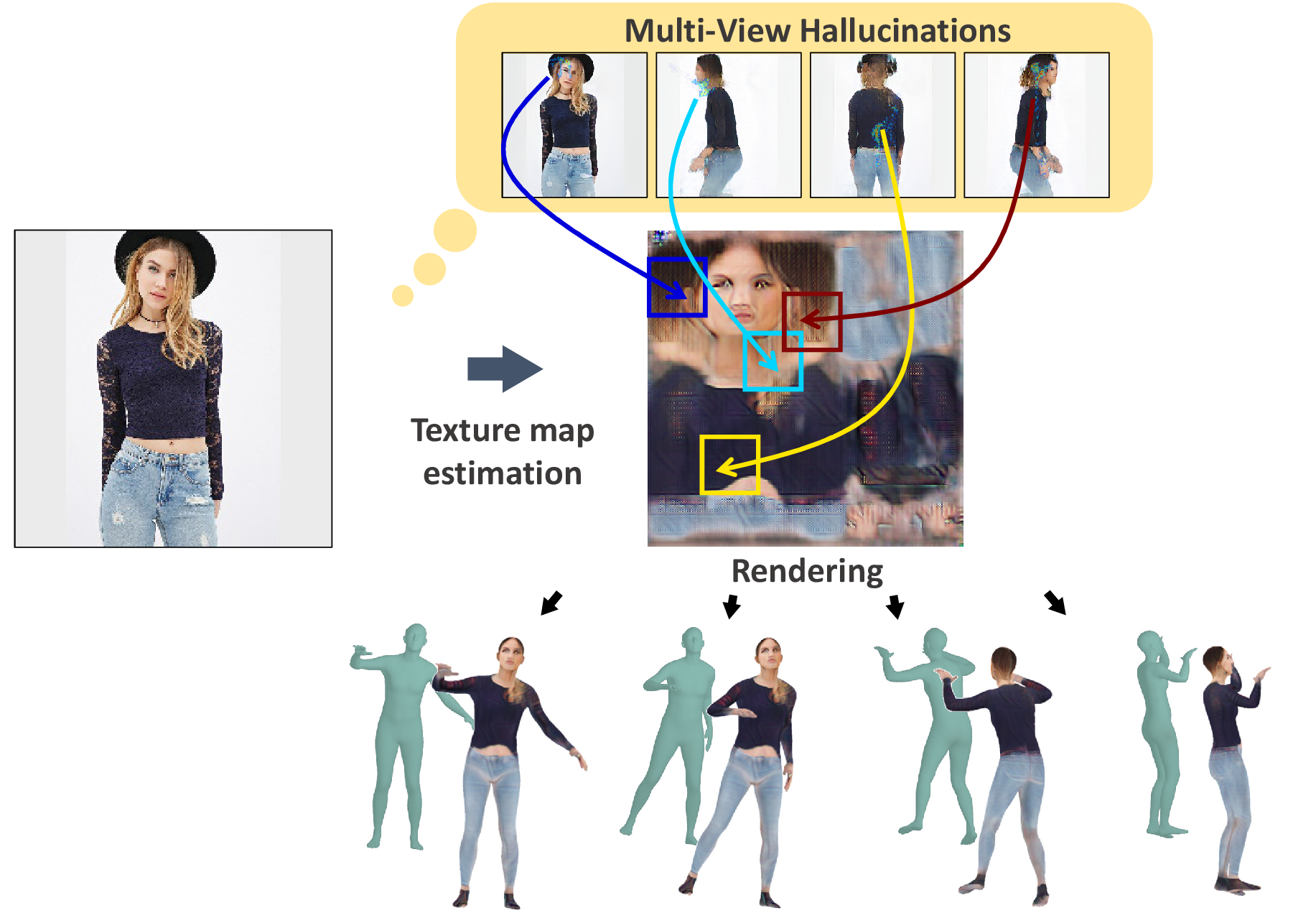}
    \caption{Estimating a texture map from a single image by hallucinating multi-view images. An arrow from each hallucination image represents a reference part for estimating a boxed region on the estimated texture map.}
    \label{fig:intro}
\end{figure}

A texture map is an image lying on $uv$-coordinates containing whole surface colors of a 3D model \citep{catmull1974subdivision,catmull19803}. When rendering an image using the 3D model, the surface color values are drawn from the texture map according to the pre-defined $uv$ parameterization. One hardness of the texture map estimation from a single image is that a single image is insufficient for generating a complete texture map for the entire surface. Conventionally, a texture map is obtained by aggregating multiple partial texture maps, obtained by unwrapping multi-view images into $uv$-coordinates, called stitching \citep{wang2001optimal,thormahlen20083d}. However, when it comes to generating a texture map from a single image, a half of the entire surface is unseen, whose colors should be filled by imagination.

To resolve this issue, we propose a texture map estimation method from a single image for colorizing a 3D human model utilizing multi-view features. Recently, deep architectures have achieved remarkable success on pose transferred image generation, and we expect that deep architectures can provide sufficient alternatives for real multi-view images. To this end, instead of generating a texture map solely, we design our model to generate images in two different domains: pose transferred images in the image domain and a texture map in the texture domain. Our model consists of two main generators: an image generator and a texture generator. During the generation process, the image generator generates pseudo multi-view images from a single image to provide an entire surface features for generating a texture map. Simultaneously, the texture generator generates the texture map utilizing features from the pseudo multi-view images, which we named as hallucinations. The overall texture generating scheme is depicted in Figure \ref{fig:intro}. The texture domain generator utilizes multi-view features drawn from the image generator and makes the image generator utilize texture features drawn from the texture domain generator using an attention mechanism. As a consequence, each domain's generator takes advantage of geometric clues and information about unseen surfaces. In light of the image domain generator, texture features provide geometrically consistent patterns over any pose, viewpoint, and scale, increasing generalization performance even for the out-of-distribution cases. From the texture domain generator's standpoint, image features provide natural colors and pattern information dedicated to generating real-looking images and clues for unseen surfaces.

To validate our method, we conduct experiments on various datasets and show the superiority of our method for generating a texture map and pose transferred images. We also demonstrate our resulting texture map can be applied to 3D human model for rendering a 3D animation clip.

We can summarize our contributions into three folds.
\begin{itemize}
    \item We propose a novel multi-view hallucination generation scheme to provide pseudo multi-view images for estimating a texture map from a single image.
    \item We propose dual-domain generators in which each domain feature is interacting with the other by an attention mechanism, which improves generated image qualities for image pose transfer and texture map estimation tasks.
    \item We generate a directly render-able texture map in decent quality for the 3D human model from a single image.
\end{itemize}

\section{Related Work}

\begin{figure*}[t]
    \centering
    \includegraphics[width=0.93\textwidth]{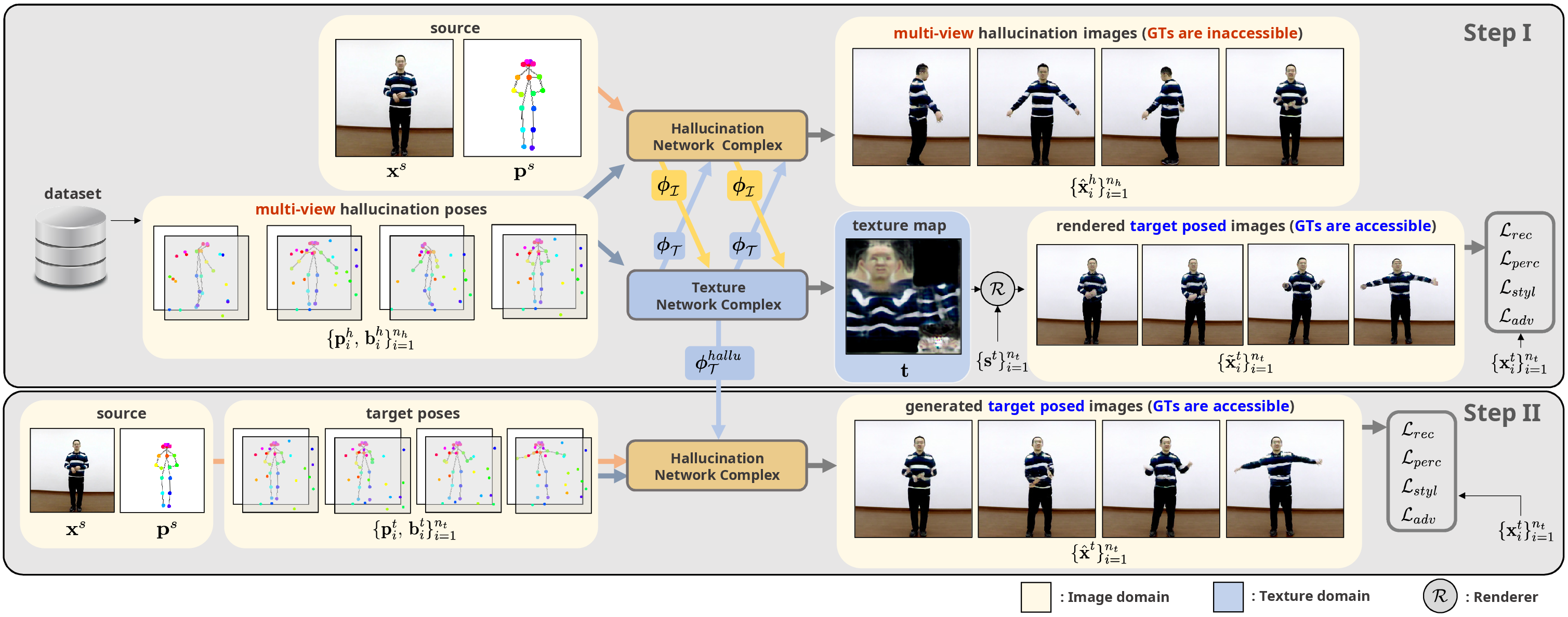}
    \caption{Overview of the proposed dual-domain generative method consisting of two generative pipelines: a \textit{hallucination network complex ($\text{H-Nets}_{\cI}$)} and a \textit{texture network complex ($\text{T-Nets}_{\cT}$)}. The $\text{H-Nets}_{\cI}$ generates $n_h$ pose-transferred hallucination images, $\{\hat{\x}^h_i\}_{i=1}^{n_h}$, on the image domain while the $\text{T-Nets}_{\cI}$ generates an estimated texture map, $\t$, on the texture domain. The two pipelines processed simultaneously exchanging their features ($\bphi_{\cI}$: image feature, $\bphi_{\cT}$: texture feature) to the other domain.}
    \label{fig:overview}
\end{figure*}

\noindent {\bf Texture map estimation.} \quad  Densepose transfer \citep{neverova2018dense}, which firstly adopts a texture mapping technique on pose transferred image generation, inpaints a partial texture map warped from an input image to generate a 
complete texture map. \citet{grigorev2018coordinate} estimate a flow-field from a partial flow map by which an input image is warped to make a full texture map. However, their resulting texture maps have limited quality for being directly used for rendering, which requires additional processing layers to obtain a final pose transferred image. \citet{wang2019reidsupervised} utilize person re-identification loss to generate a direct render-able texture map, albeit their results are somewhat blurry. \citet{lazova2019360} proposed texture and displacement-map-generating networks from a single image trained using full texture maps obtained from elaborately synthesized 3D models. The generated texture map by \citet{lazova2019360} has sufficient fidelity, however, training them requires full texture maps which are hard to obtain in practice.
Utilizing texture mapping for pose transferred image generation has an advantage of keeping temporal consistency on the same surfaces when generating a video clip. \citet{zhi2020texmesh} proposed a texture and displacement generating framework from multiple RGB-D frames of a video. Our approach differs from \citet{zhi2020texmesh} in that we assume multi-view images for generating a texture map are not given as inputs, but an another objective to generate during the process.

\section{Preliminary}
\noindent \textbf{Global flow local attention (GFLA) \citep{ren2020deep}} is a patch-based attention module in which a local patch is extracted from where a flow points to. GFLA consists of two modules: \textit{flow generator,}\footnote{It is called a \textit{global flow field estimator} in the original paper.} $F$, and \textit{local attention module}, $A$. A flow generator $F$ generates a flow $\f$, according to which local patches are extracted\footnote{Originally, $\f$ represents relative positions, however, we use relative positional representation when key and query features lie on the same domain, and absolute positional representation elsewhere.}, and a binary mask $\m$ for merging features. Let $\bphi^q$ and $\bphi^k$ denote a query and a key feature respectively, and $\f^{kq}$ denote a flow from key to query. The local attention module outputs $\bphi^{out} = A(\bphi^q, \bphi^k, \f^{kq})$ in two steps. Let $\cN_n(\bphi,\,l)$ be an $n\times n$ sized local patch extracted from $\bphi$ centered at location $l$. In a local attention module, a local attention feature $\bphi^{attn}$ is computed as
\begin{equation}
    \bphi^{attn}(l) = \text{Attn}\Big(\cN_n(\bphi^q,\,l),\,\cN_n\big(\bphi^k,\,l+\f^{kq}(l)\big)\Big),
    \label{eq:attention}
\end{equation}
where $\text{Attn}(\cdot,\,\cdot)$ is a general attention module \citep{vaswani2017attention}. Then the final output $\bphi^{out}$ is computed as
\begin{equation}
    \bphi^{out} = (\one - \m^{kq}) \otimes \bphi^q + \m^{kq} \otimes \bphi^{attn}.
    \label{eq:attention_out}
\end{equation}
where $\m^{kq}$ denotes a binary mask generated by $F$ together with $\f^{kq}$, $\otimes$ denotes an element-wise multiplication, and $\one$ denotes a tensor whose elements are all ones.

\section{Proposed Method}


\begin{figure*}[t]
    \centering
    \includegraphics[width=0.9\textwidth]{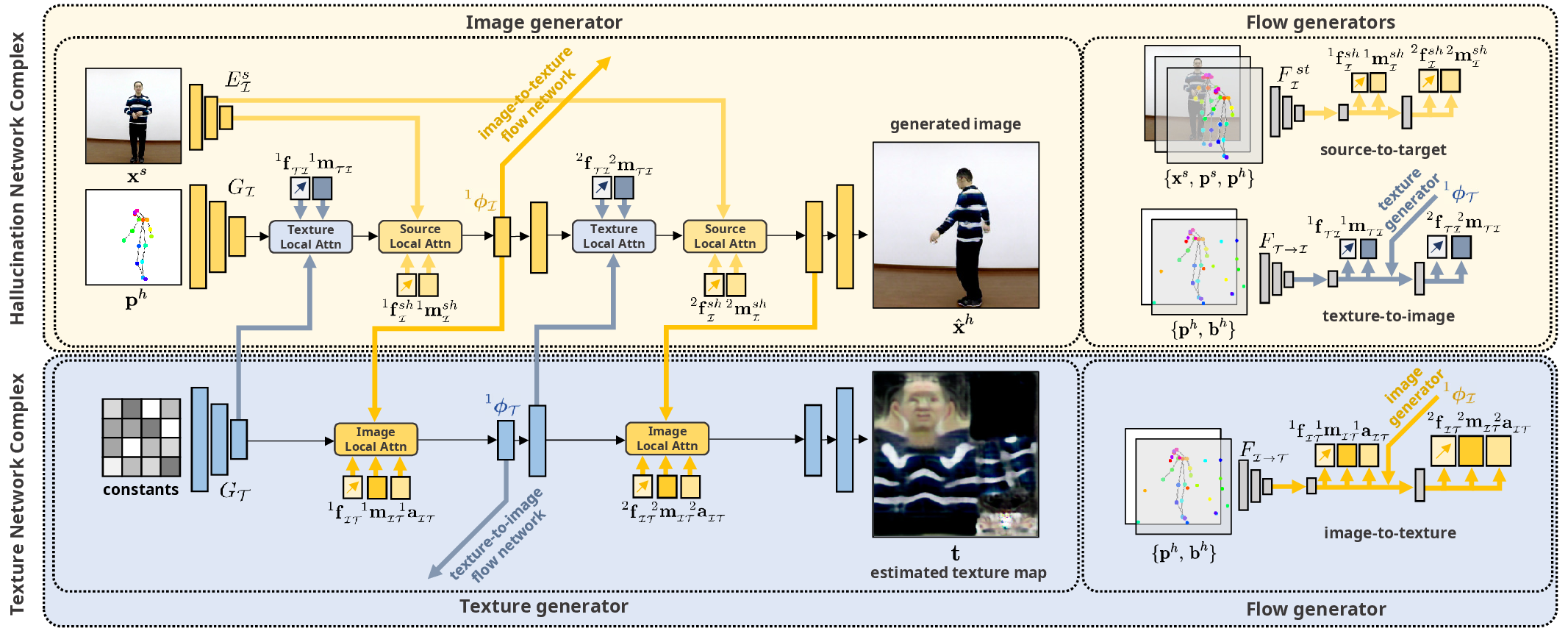}
    \caption{Detailed structures of the hallunet-complex and the texnet-complex consisting of a respective domain generator and the corresponding flow generator(s) with input-output linkages.}
    \label{fig:network_structure}
\end{figure*}

\noindent \textbf{Notations.} \quad Let $\x$ denote an image, $\s$ denote a surface annotation representing texel\footnote{pixel of a texture map} coordinates of pixels in $uv$, obtained by DensePose \citep{guler2018densepose}. Let $\p$ denote an image pose of the image $\x$ represented as a heat map of keypoints detected by OpenPose \citep{cao2019openpose}. Let $\t$ denote an estimated texture map and $\c$ denote a coordinate annotation representing pixel coordinates of texels. The coordinate annotation $\c$ and the surface annotation $\s$ are inversely related satisfying $l=\c(\s(l))$ for any pixel coordinate $l$ on a human body. Let $\b$ denote a texture pose, a warped image pose $\p$ to the texture domain according to the coordinate annotation $\c$, namely $\b = warp(\p;\c)$. Superscript $s$ and $t$ are used to denote source and target, identifying that a symbol is used for pre-/post-pose-transform, respectively, and $h$ is also used in place of $t$ to emphasize that targets are used for hallucination.
Please refer to the supplementary material for more detailed notations.

\noindent \textbf{Formulation.} \quad Our model consists of two generative network pipelines: a \emph{hallucination network complex} (\emph{hallunet-complex}, $\text{H-Nets}_{\cI}$) and a \emph{texture network complex} (\emph{texnet-complex}, $\text{T-Nets}_{\cT}$). Step I in Figure \ref{fig:overview} depicts an overview of the generation process. The hallunet-complex, $\text{H-Nets}_{\cI}$, simultaneously generates $n_h$ pose transferred hallucination images, $\{\hat{\x}^h_i\}_{i=1}^{n_h}$, from a source image, $\x^s$, and $n_h$ hallucination image poses, $\{\p^h_i \}_{i=1}^{n_h}$, while the texnet-complex, $\text{T-Nets}_{\cT}$, generates an estimated texture map, $\t$. The hallunet-complex and the texnet-complex are processed simultaneously referring to intermediate features of the other domain by GFLA. To utilize GFLA, the hallunet-complex, $\text{H-Nets}_{\cI}$, consists of a texture-to-image flow generator admitting hallucination image poses, $\{\p^h_i\}_{i=1}^{n_h}$, as inputs. Similarly, the texnet-complex, $\text{T-Nets}_{\cT}$, consists of a image-to-texture flow generator admitting hallucination poses on both image and texture domain, $\{\p^h_i, \b^h_i\}_{i=1}^{n_h}$, as inputs. Denoting $\bphi_{\cI}$ and $\bphi_{\cT}$, intermediate features\footnote{The actual attention mechanism is applied to multiple feature layers, however, we regard them as a single layer feature for a concise representation in the rest of the paper.} of the image domain and the texture domain, respectively, we can express a generation process of each network complex as
\begin{gather}
    \hat{\x}^h_i,\,{\bphi^h_{\cI}}_i = \text{H-Nets}_{\cI}(\x^s,\,\p^s,\,\p^h_i,\,\bphi_{\cT}),\\
    \t,\,\bphi_{\cT} = \text{T-Nets}_{\cT}(\P^h,\,\B^h,\,\bPhi^h_{\cI}),
    \label{eq:texnet_complex}
\end{gather}
where a capital symbol denotes a set of $n_h$ smaller symbols used for hallucinations, \textit{e.g.}, $\P^h = \{\p^h_i\}_{i=1}^{n_h}$ and $\bPhi^h_{\cI}=\{{\bphi^h_{\cI}}_i\}_{i=1}^{n_h}$.

\subsection{Hallucination Network Complex}

A hallucination network complex consists of a source image encoder, $E_{\cI}^s$, an \textit{image generator}, $G_{\cI}$, and two flow generators: \textit{source-to-target (source-to-hallucination) flow generator}, $F_{\cI}^{st}$, and \textit{texture-to-image flow generator}, $F_{\cT\rightarrow\cI}^{\,}$.


\noindent \textbf{Image generator.} \quad The image generator, $G_{\cI}$, generates $n_h$ pose transferred hallucination images, $\{\hat{\x}^h_i\}_i^{n_h}$, each of which is generated from a source image, $\x^s$, conditioned on a hallucination image pose, $\p^h_i$. The image generator consists of an encoder-decoder structure interleaved with two types of local attention modules -- \textit{source local attention module}, $A_{\cI}^{st}$, and \textit{texture local attention module}, $A_{\cT\rightarrow\cI}$, -- at decoder side. The source image encoder $E_{\cI}^s$ provides a source image feature extracted from the source image $\x^s$ as a key feature for the source local attention module. Let $\bphi^s_{\cI}$ denote a source image feature, $\bphi_{\cT}$ denote a texture feature, $\f^{\,sh}_{\cI}$ and $\m^{sh}_{\cI}$ denote a source-to-hallucination flow and a corresponding mask respectively. Let $\f_{\cT\cI}$ and $\m_{\cT\cI}$ denote a texture-to-image flow and a corresponding mask. Then the image generator $G_{\cI}$ generates $n_t$ target images $\{\hat{\x}^h_i\}_{i=1}^{n_h}$ as
\begin{equation}
    \hat{\x}^h_i = G_{\cI}(\p^h_i, \bphi^s_{\cI}, \bphi_{\cT}, \f_{\cI}^{\,sh}, \m_{\cI}^{sh}, \f_{\cT\cI}, \m_{\cT\cI}),
\end{equation}
referring to the source feature, $\bphi_{\cI}^s$, and the texture feature, $\bphi_{\cT}$, as a query using a \textit{source-} and a \textit{texture-} local attention module by \eqref{eq:attention} and \eqref{eq:attention_out} respectively.


\noindent \textbf{Source-to-target (source-to-hallucination) flow generator.} \quad The source-to-target flow generator, $F_{\cI}^{st}$, generates a source-to-hallucination flow $\f^{\,sh}_{\cI}$ and a corresponding mask $\m^{sh}_{\cI}$ for the source local attention module from the source image $\x^s$, the source pose $\p^s$ and the hallucination image pose $\p^h$, following the GFLA \citep{ren2020deep}, as
\begin{equation}
    \f^{\,sh}_{\cI},\,\m^{sh}_{\cI} = F_{\cI}^{\,sh}(\x^s,\,\p^s,\,\p^h).
\end{equation}
\noindent \textbf{Texture-to-image flow generator.} \quad The texture-to-image flow generator, $F_{\cT\rightarrow\cI}$, generates a texture-to-image mask, $\m_{\cT\cI}$, for texture a local attention module basically from a hallucination image pose, $\p^h$, for the lowest layer and sequentially combines a texture feature, $\bphi_{\cT}$, of the same level layer after outputting the lowest layer's flow and mask, as depicted in Figure \ref{fig:network_structure}.
\begin{equation}
    \f_{\cT\cI},\, \m_{\cT\cI} = F_{\cT\rightarrow\cI}(\bphi_{\cT},\,\p^h).
\end{equation}

\subsection{Texture Network Complex}

A texture network complex consists of a \textit{texture generator}, $G_{\cT}$, and a \textit{image-to-texture flow generator}, $F_{\cI\rightarrow\cT}$.

\noindent \textbf{Texture generator.} \quad The texture generator, $G_{\cT}$, generates an estimated texture map, $\t$, from learn-able constants utilizing multi-pose image features, $\bPhi^t=\{\bphi^t\}_{i=1}^{n_t}$. The SMPL \citep{loper2015smpl} based texture map we use has a specific layout, in which every body part appears at the same location. Hence, we design the learn-able constants be the universal input for the texture generator, $G_{\cT}$, to generate all output texture maps, hoping the texture generator to find an optimal encoding for the universal texture map structure. Similar to the image generator, the texture generator is composed of an encoder-decoder structure interleaved with image local attention modules at decoder side. While the image generator, $G_{\cI}$, generates $n_t$ different target images simultaneously, the texture generator, $G_{\cT}$, generates a single texture map, $\t$, referring to $n_t$ different image features at once. Let $\bPhi_{\cI} = \{\bphi_{{\cI}_i}\}_{i=1}^{n_t}$ denote a set $n_t$ image features of $n_t$ different target poses generated by the image generator, and $\F_{\cI\cT} = \{\f_{{\cI\cT}_i}\}_{i=1}^{n_t}$, $\M_{\cI\cT}=\{\m_{{\cI\cT}_i}\}_{i=1}^{n_t}$ denote a set of $n_t$ image-to-texture flows and masks, and $\a_{\cI\cT}$ denote an aggregation mask which will be explained later. Then the texture generator, $G_{\cT}$, estimates a texture map, $\t$, as
\begin{equation}
    \t = G_{\cT}(\bPhi_{\cI}, \F_{\cI\cT}, \M_{\cI\cT}, \a_{\cI\cT}).
    \label{eq:texture_generator}
\end{equation}

We design the texture generator to attend to multi-view features at the same time. To achieve this, we introduce an additional merging layer at the end of the image local attention module to merge multiple multi-view attention features. Let $n_h$ be the number of hallucination and $\bphi^{attn}_i$ be the attention feature of $i$-th view according to \eqref{eq:attention}. Then a merged attention feature, $\bphi_{merge}$, is obtained by applying a convolution layer on a concatenation of $\{\bphi^{attn}_i\}_{i=1}^{n_h}$ as
\begin{equation}
    \bphi^{merge} = \text{Conv}(\text{concat}(\bphi^{attn}_1\,,\,...\,,\,\bphi^{attn}_{n_h})),
\end{equation}
where $\text{concat}(\cdot)$ denotes a concatenation operation on features along the channel dimension. Then the final output feature, $\bphi^{out}$, is computed as
\begin{equation}
    \bphi^{out} = (\one - \a_{\cI\cT}) \otimes \bphi_{\cI\cT} + \a_{\cI\cT} \otimes \bphi^{merge},
    \label{eq:merge_attention}
\end{equation}
with a texture decoding feature, $\bphi_{\cT}$, and an aggregation mask, $\a_{\cI\cT}$, obtained by $F_{\cI\rightarrow\cT}$, where $\one$ is a tensor whose elements are all ones.

\noindent \textbf{Image-to-texture flow generator.} \quad The image-to-texture flow generator, $F_{\cI\rightarrow\cT}$, generates an image-to-texture flow, $\f_{\cI\cT}$, a corresponding mask, $\m_{\cI\cT}$, and an aggregation mask, $\a_{\cI\cT}$, for image local attention from a hallucination image pose, $\p^h$, a hallucination texture pose, $\b^h$, and a hallucination pixel coordinate, $\c^h$, as
\begin{equation}
    \f_{\cI\cT},\,\m_{\cI\cT},\,\a_{\cI\cT} = F_{\cI\rightarrow\cT}(\p^h,\,\b^h,\,\c^h).
\end{equation}
Notice that the $F_{\cI\rightarrow\cT}$ generates the additional aggregation mask $\a_{\cI\cT}$ for merging multi-view attention features according to \eqref{eq:merge_attention}.

\subsection{Loss Functions and Training Strategy}

We assume that the ground-truth of an estimated texture map is inaccessible. Hence, we render an image using the estimated texture map, $\t$, and compare it to a ground-truth image for training. We simplify the rendering process into warping $\t$ according to a surface coordinate, $\s$. However, the warped texture map constitutes only a foreground human body and lacks background. We provide the lacking background to the rendered image from the generated pose transferred image, $\hat{\x}$, generated by the image generator. Let $\m$ denote a binary mask of the surface coordinate, $\s$. Then we obtain the final rendered image, $\tilde{\x}$, by
\begin{equation}
    \tilde{\x} = (\one - \m) \otimes \hat{\x} + \m \otimes warp(\t;\,\s).
    \label{eq:rendering}
\end{equation}


\begin{figure}[t]
    \centering
    \includegraphics[width=0.6\columnwidth]{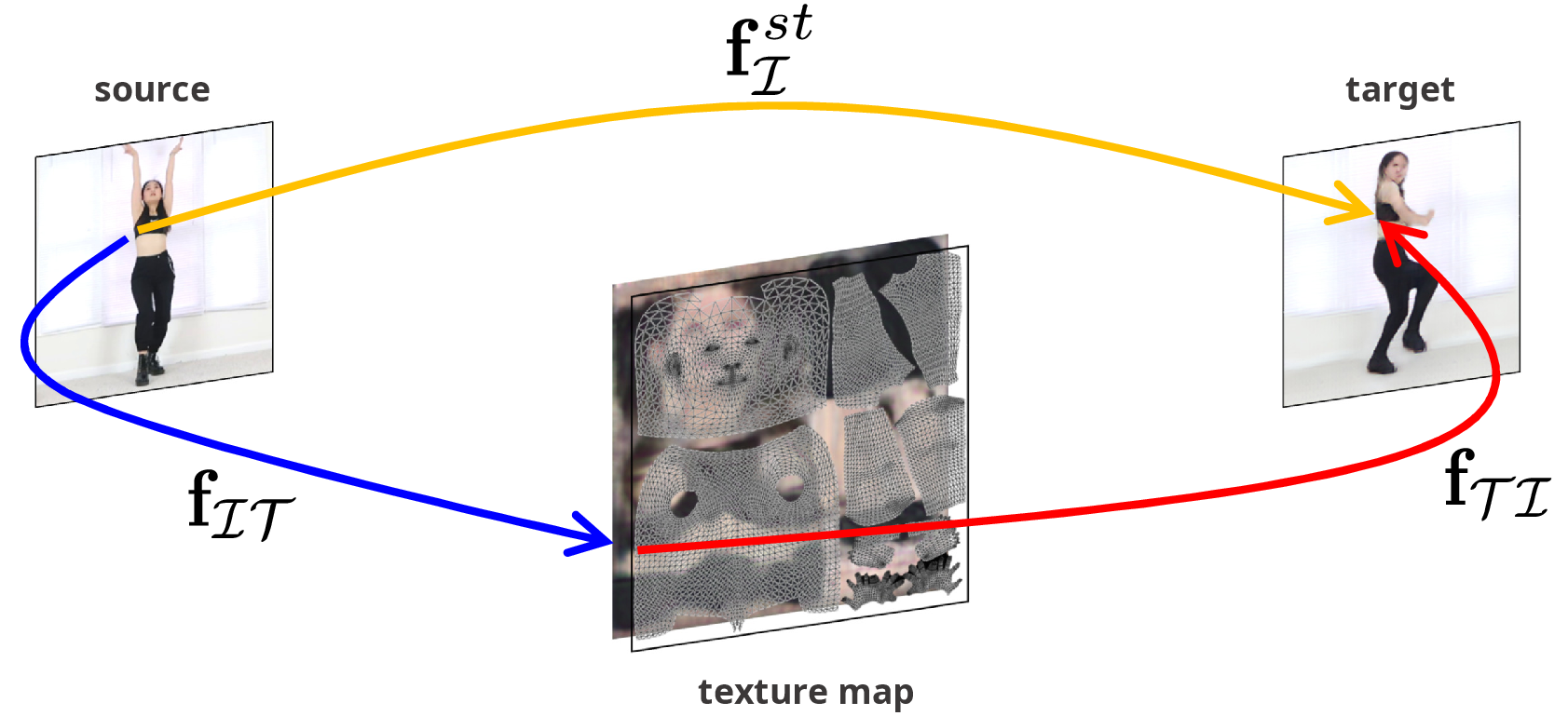}
    \caption{A path along the combined flows, $\f_{\cI\cT}$ and $\f_{\cT\cI}$, should be consistent with a direct path along the $\f_{\cI}^{\,st}$.}
    \label{fig:path_consistency}
\end{figure}

\noindent \textbf{Loss functions.} \quad To train the image generator, $G_{\cI}$, and the texture generator, $G_{\cT}$, we use four types of losses in the image domain: Reconstruction loss, $\cL_{rec}$, to minimize the difference between a generated/rendered image, $\{\hat{\x},\tilde{\x}\}$, and ground-truth image, $\x$, according to $\ell1$ norm as
\begin{equation*}
    \cL_{rec} = \| \hat{\x} - \x \|_1 + \| \tilde{\x} - \x \|_1. 
\end{equation*}
The perceptual loss, $\cL_{perc}$ \citep{johnson2016perceptual}, to minimize $\ell1$ norm between deep features of $\{\hat{\x},\tilde{\x}\}$ and $\x$ as
\begin{equation*}
    \cL_{perc} = \sum_j \big[\, \| \prescript{j}{}{\bpsi}(\hat{\x}) -  \prescript{j}{}{\bpsi}(\x) \|_1 + \| \prescript{j}{}{\bpsi}(\tilde{\x}) - \bpsi^j(\x) \|_1 \big],
\end{equation*}
where $\prescript{j}{}{\bpsi}$ represents a $j$-th layer feature obtained by the pre-trained VGG19 networks for preserving coarse level contents. The style loss, $\cL_{styl}$ \citep{johnson2016perceptual},
\begin{equation*}
    \cL_{styl} = \sum_j \big[\, \| \prescript{j}{}{G_{\psi}}(\hat{\x}) - \prescript{j}{}{G_{\psi}}(\x) \|_1 + \| \prescript{j}{}{G_{\psi}}(\tilde{\x}) - \prescript{j}{}{G_{\psi}}(\x) \|_1 \big],
\end{equation*}
for preserving an overall style, where $\prescript{j}{}{G_{\bpsi}}$ represents a Gram matrix constructed from $\prescript{j}{}{\bpsi}$. And the hinge version of the adversarial loss, $\cL_{adv}$, with a discriminator, $D(\cdot)$, to make the generated/rendered images and the estimated texture map real-looking.

Additionally, we use four types of losses to train three flow generators: $F_{\cI}^{\,st}$, $F_{\cT\rightarrow\cI}$, and $F_{\cI\rightarrow\cT}$. As in \citet{ren2020deep}, we use the sample correctness loss to train the source-to-target flow generator, $F_{\cI}^{\,st}$:
\begin{equation}
    \cL_{cor} = \frac{1}{L} \sum_{l\in \Omega} \exp\bigg(-\frac{\mu(\tilde{\bpsi}_{s}^l,\, \bpsi_t^l)}{\mu_{max}^l}\bigg),
\end{equation}
where $\mu(\cdot, \cdot)$ denotes the cosine similarity, $\Omega$ denotes the coordinate set containing all $L$ positions in the feature maps, $\tilde{\bpsi}_{s}$ denotes the warping of the VGG19 feature, $\bpsi_s$, according to the flow $\f_{\cI}^{\,st}$, that is $\tilde{\bpsi}_{s}=warp(\bpsi_s;\,\f_{\cI}^{\,st})$, with a superscript $l$ denoting feature values of $\tilde{\bpsi}_{s}$ located at the coordinate $l=(x,y)$.
To train the image-to-texture flow generator, $F_{\cT\cI}$, we introduce a coordinate loss, $\cL_{coord}$, as
\begin{equation}
    \cL_{coord} = \big\|\tilde{\m}_{\cT} \otimes (\f_{\cI\cT} - \tilde{\c}) \big\|_2,
    \label{eq:loss_coord}
\end{equation}
where $\tilde{\c}$ denotes a rescaled version of $\c$ to the same spatial size and scale of $\c_{\cI\cT}$, $\tilde{\m}_{\cT}$ denotes a binary mask indicating visible parts of $\tilde{\c}$. Additionally, we introduce a path consistency loss, $\cL_{cons}$. Considering two types of paths as depicted in Figure \ref{fig:path_consistency}. One path, represented as $\f_{\cI}^{\,st}$, is a direct path from a source to a target. The other path is a two-step path from the source to the target passing through a texture map represented as a combination of image-to-texture flow, $\f_{\cI\cT}$, and texture-to-image flow, $\f_{\cT\cI}$. We assume that information contained in the source image should be convey to the same location on the target image regardless of the paths. To impose this assumption, the path consistency loss reduces the difference between the two paths as
\begin{equation}
    \cL_{cons} = \big\| \m \otimes \big( \f_{\cI}^{\,st} - warp(\f_{\cI\cT};\,\f_{\cT\cI}) \big)  \big\|_2,
\end{equation}
with the binary mask, $\m$, representing foreground human body obtained along with surface annotation, $\s$. Lastly, all flows are regularized by the regularization loss devised in \citet{ren2020deep} as 

\begin{equation}
    \mathcal{L}_{reg} = \mathcal{L}_r (\f_{\cI}^{\,st}) + \mathcal{L}_r (\f_{\cI\cT}).
\end{equation}
Please refer to \citet{ren2020deep} for further details of the regularization loss.

\noindent \textbf{Training Strategy} \quad The goal of the hallunet-complex is to provide sufficient image features from diverse viewpoints to the texture generator. Providing evenly rotated poses as a set of hallucination poses could be an option, however, generating evenly rotated images is often ungeneralizable for the image generator as most training images are biased to frontal and side views. To balance the trade-off between viewpoint diversity and generalization performance, we sample $n_h \unaryminus 1$ poses from 
another image pair having a different clothes identity and combine a source pose to make a set of $n_h$ hallucination poses. Let $\{\p_i^h\}_{i=1}^{n_h}$ denote a set of hallucination image poses. As $\{\p_i^h\}_{i=1}^{n_h}$ are sampled from the other image pair, except one from the source, we do not have ground-truths to evaluate the generated hallucination images. Hence we propose two-step generation processes for training as depicted in Figure \ref{fig:overview}. Firstly, we run the whole networks, both hallunet-complex and texnet-complex, using the sampled hallucination poses, $\{\p_i^h\}_{i=1}^{n_h}$, to obtain an estimated texture map, $\t$, and a texture feature, $\bphi_{\cT}$, which we named it $\bphi_{\cT}^{hallu}$. Let $\{\p_i^t\}_{i=1}^{n_t}$ denote target image poses of the current image pair, a set of different pose images of the source image which we can use as ground-truths. In the second step, we run the hallunet-complex solely using the target image poses, $\{\p_i^t\}_{i=1}^{n_t}$, referring to the kept hallucination texture feature, $\bphi_{\cT}^{hallu}$, to obtain generated target images, $\{\hat{\x}_i^t\}_{i=1}^{n_t}$, posing $\{\p_i^t\}_{i=1}^{n_t}$. Now, we do have the ground-truths for $\{\hat{\x}_i^t\}_{i=1}^{n_t}$, we can train the whole networks using the proposed loss functions.


\section{Experiments}

\noindent \textbf{Datasets.} \quad We use three datasets to evaluate our model: DeepFahsion In-shop Clothes Retrieval Benchmark \cite{liu2016deepfashion}, iPER \citep{liu2019liquid}, and Fashion video collected from Amazon \citep{zablotskaia2019dwnet}. From DeepFahsion we filter out 5,745 images wearing 1,628 different clothes, which are non-detectable to the human detector \citep{cao2019openpose}, from the training set.

\noindent \textbf{Evaluation and metrics.} \quad To evaluate estimated texture maps, we render multi-pose/view images using the estimated texture maps as we do not have ground-truth texture maps. We use three measures to compute reconstruction errors and a distributional discrepancy between generated images and reference images: Structural similarity (SSIM) \citep{wang2004image}, Learned Perceptual Image Patch Similarity (LPIPS) \citep{zhang2018unreasonable}, and Fr\'echet inception distance (FID) \citep{heusel2017gans}.

\subsection{Comparisons}

\begin{figure}[t]
    \centering
    \includegraphics[width=\columnwidth]{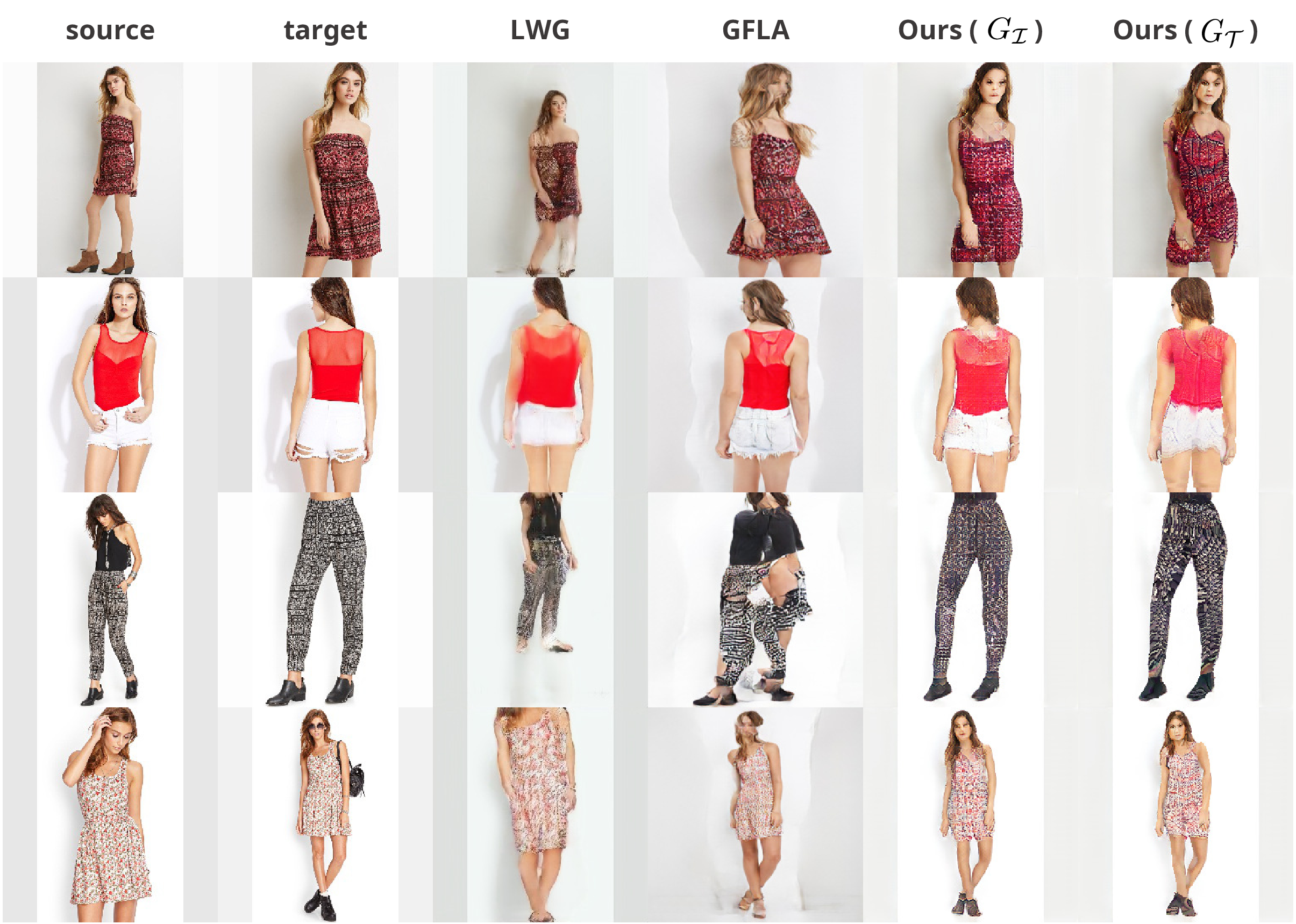}
    \caption{Examples of comparison result on Deepfashion dataset to pose transferred image generation methods. LWG \citep{liu2019liquid} and GFLA \citep{ren2020deep} preserve textures locally, but sometimes fail to generate exact posture and scaled images, while ours preserve the overall postures and scales.}
    \label{fig:comparison}
\end{figure}

\begin{table}[t]
\begin{center}
\caption{Comparison results on DeepFashion and iPER datasets.
The image domain output (Ours ($G_{\cI}$)), and the rendered outputs using the texture domain output (Ours ($G_{\cT}$)) are compared.}
\label{tab:comparison}
\begin{adjustbox}{max width=0.9\columnwidth}
\begin{tabular}{l||*{3}{c}|*{2}{c}}
\hline
\multirow{2}{*}{ } & \multicolumn{3}{c|}{DeepFashion} & \multicolumn{2}{c}{iPER} \\
\cline{2-6}
 & FID $\downarrow$ & LPIPS $\downarrow$ & SSIM $\uparrow$ 
 & LPIPS $\downarrow$ & SSIM $\uparrow$ \\
\hline\hline
PG2 & 47.714 & 0.246 & 0.763 
    & 0.135 & 0.854 \\
Def-GAN & 18.457 & 0.233 & 0.761
        & 0.129 & 0.829 \\
LWG & 23.286 & 0.283 & 0.731 
    & 0.087 & 0.840 \\
GFLA & 10.573 & 0.234 & 0.715
     & - & - \\
HPBTT & - & - & 0.735
      & - & - \\
\hline
Ours ($G_{\cI}$) & \textbf{9.001} & \textbf{0.156} & \textbf{0.830}
         & \textbf{0.051} & \textbf{0.907} \\
Ours ($G_{\cT}$) & 19.656 & 0.177 & 0.786
         & 0.063 & 0.897 \\
\hline
\end{tabular}
\end{adjustbox}
\end{center}
\end{table}

We compare our method to several state-of-the-art pose guided image transfer methods including PG2 \citep{ma2017pose}, Def-GAN \citep{siarohin2018deformable}, GFLA \citep{ren2020deep}, LWG \citep{liu2019liquid}, and a recent texture map estimation method, HPBTT \citep{zhao2020human}. The results are summarized in Table \ref{tab:comparison}. On the DeepFashion and iPER dataset our method outperforms the others. Figure \ref{fig:comparison} shows some examples comparing generated images (ours ($G_{\cI}$)) and rendered images (ours ($G_{\cT}$)) of our method to other pose transferred image generation methods. We can find that our model has the advantage of generating an image involving large scale transform over the others. For example, in the third row of Figure \ref{fig:comparison}, LWG and GFLA generate fine details locally but fail at generating the exact pose and consistent patterns overall. However, our method successfully generates a desirable scaled image even if the target pose represents merely a magnified body part, ascribing to the interacting feature flow. The rendering results of ours are comparable to the others despite some artifacts, attributing to resolution mismatch between the image and the texture map and surface annotation errors. Thus, we can conclude that our texture generator generates a plausible texture map for direct rendering. Please refer to the supplementary material for more examples.



\begin{figure*}[t]
    \centering
    \includegraphics[width=\textwidth]{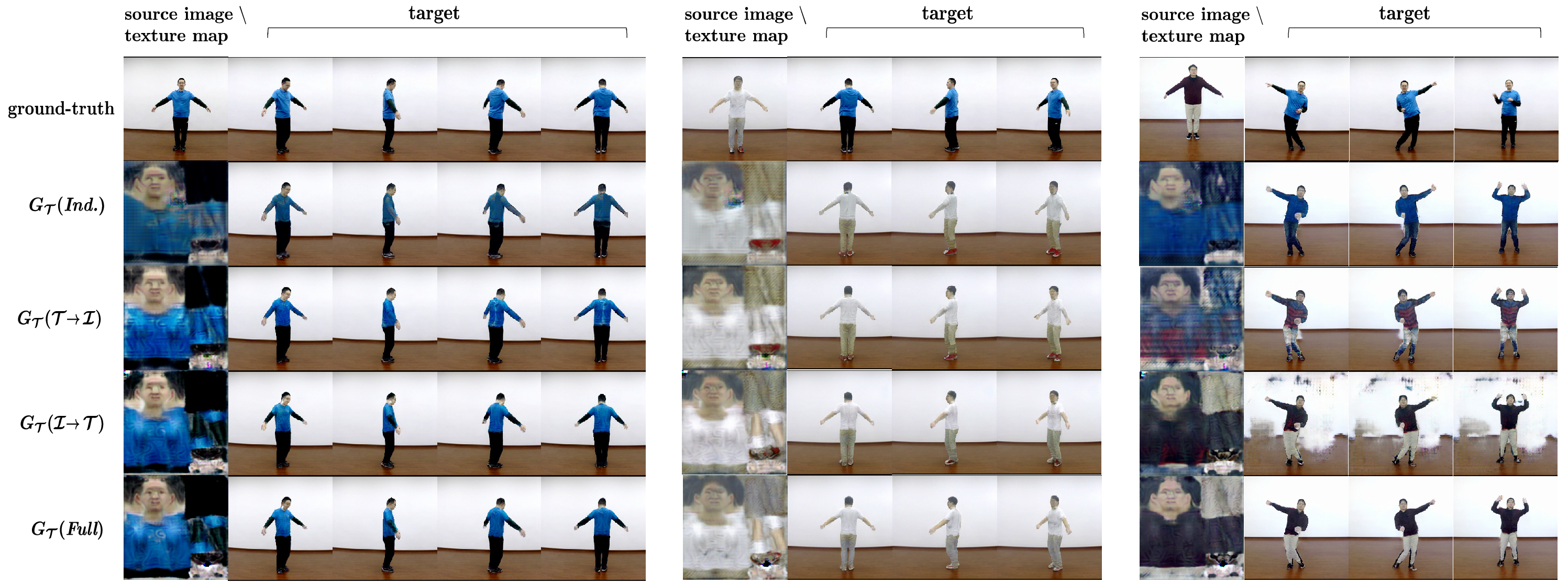}
    \caption{Examples of ablation studies. Rendered images of various viewpoints and poses using a estimated texture map from a source image (column 1 of each example, top: source image, rest: texture map) of \textit{Independent models} (row 2), \textit{Texture-to-image models} (row 3), \textit{Image-to-texture models} (row 4), and the \textit{Full models} (row 5).}
    \label{fig:ablation}
\end{figure*}

\subsection{Analysis and Ablation Study}
\label{sec:ablation}

\begin{table}[t]
    \begin{center}
    \caption{Analysis results of the number of hallucination.}
    \label{tab:analysis_num_hallu}
    \begin{adjustbox}{max width=0.85\columnwidth}
    \begin{tabular}{l||*{2}{c}|*{2}{c}}
        \hline
        \multirow{2}{*}{} & \multicolumn{2}{c|}{iPER} & \multicolumn{2}{c}{FashionVid}\\
        \cline{2-5}
        & LPIPS $\downarrow$ & SSIM $\uparrow$ & LPIPS $\downarrow$ & SSIM $\uparrow$\\
        \hline
        \hline
         $G_{\cI}$ ($n_h = 1$) & 0.067 & 0.894 & 0.063 & 0.920 \\
         $G_{\cI}$ ($n_h = 2$) & 0.055 & \textbf{0.907} & \textbf{0.061} & 0.922 \\
         $G_{\cI}$ ($n_h = 3$) & \textbf{0.051} & \textbf{0.907} & 0.066 & 0.921 \\
         $G_{\cI}$ ($n_h = 4$) & 0.082 & 0.871 & \textbf{0.061} & \textbf{0.924} \\
         \hline
         $G_{\cT}$ ($n_h = 1$) & 0.076 & 0.887 & 0.081 & 0.903 \\
         $G_{\cT}$ ($n_h = 2$) & 0.066 & \textbf{0.898} & 0.081 & \textbf{0.905} \\
         $G_{\cT}$ ($n_h = 3$) & 0.063 & 0.897 & 0.082 & 0.903 \\
         $G_{\cT}$ ($n_h = 4$) & \textbf{0.026} & 0.869 & \textbf{0.079} & \textbf{0.905} \\
         \hline
    \end{tabular}
    \end{adjustbox}
    \end{center}
\end{table}

\noindent \textbf{Number of hallucination.} \quad To analyze whether the proposed hallucination generation scheme is indeed helpful for texture map estimation, we conduct experiments increasing the number of hallucination, $n_h$, from one to four. Table \ref{tab:analysis_num_hallu} summarizes the results. For iPER, LPIPS tends to decrease for increasing $n_h$ on both generated and rendered images. In terms of SSIM, generation quality greatly increases for $n_h \shortequal 2$ compared to $n_h\shortequal 1$ for all cases, which demonstrates the effectiveness of the proposed hallucination generation scheme for texture map estimation. However, there are little improvements for $n_h > 2$ and a degenerate result appears for $G_{\cT} (n_h\shortequal 4)$. Practically, as each posed image reveals a half of the whole surface, $n_h\shortequal 2$ seems sufficiently enough to contain all surface features. We conjecture the degenerate result for $n_h\shortequal 4$ on iPER is attributed to overlapping surfaces among hallucinations which distract both image and texture generators from generating qualified outputs.

\begin{table}[t]
    \begin{center}
    \caption{Results of the ablation study. $G_{\cI}$ denote generated images from the image generator and $G_{\cT}$ denote rendering images using the estimated texture maps from the texture generator.}
    \label{tab:ablation}
    \begin{adjustbox}{max width=0.85\columnwidth}
    \begin{tabular}{l||*{2}{c}|*{2}{c}}
        \hline
        \multirow{2}{*}{} & \multicolumn{2}{c|}{iPER} & \multicolumn{2}{c}{FashionVid}\\
        \cline{2-5}
         & LPIPS $\downarrow$ & SSIM $\uparrow$
         & LPIPS $\downarrow$ & SSIM $\uparrow$\\
        \hline
        \hline
        $G_{\cI}$ (\emph{Ind.}) & \textbf{0.051} & \textbf{0.907} & 0.073 & 0.913 \\
        $G_{\cI}$ ($\cT\shortrarrow\cI$) & 0.056 & 0.903 & 0.072 & 0.916 \\
        $G_{\cI}$ ($\cI\shortrarrow\cT$) & 0.061 & 0.901 & \textbf{0.057} & 0.900 \\
        $G_{\cI}$ (\emph{Full}) & \textbf{0.051} & \textbf{0.907} & 0.066 & \textbf{0.921} \\
        \hline
        $G_{\cT}$ (\emph{Ind.}) & 0.075 & 0.888 & 0.100 & 0.888 \\
        $G_{\cT}$ ($\cT\shortrarrow\cI$) & 0.075 & 0.887 & 0.085 & 0.901 \\
        $G_{\cT}$ ($\cI\shortrarrow\cT$) & 0.073 & 0.892 & 0.083 & 0.902 \\
        $G_{\cT}$ (\emph{Full}) & \textbf{0.063} & \textbf{0.897} & \textbf{0.082} & \textbf{0.903} \\
        \hline
    \end{tabular}
    \end{adjustbox}
    \end{center}
\end{table}

\noindent \textbf{Ablation study.} \quad To analyze the role of inter-domain feature flows, we conduct ablation studies by unlinking each attention path from one domain to the other. \textit{Independent model (Ind.)} has no inter-domain attention path, \textit{Image-to-texture model ($\cI\shortrarrow\cT$)} solely has image-to-texture attention path, and \textit{Texture-to-image models ($\cT\shortrarrow\cI$)} solely has texture-to-image attention path. As the original input for the texture generator have no distinguishable information about input clothes, we leave the image-to-texture flow at the lowest layer solely from the source image to the texture generator for the \textit{Ind.} and the $\cT\shortrarrow\cI$ models. \textit{Full model (Full)} denote the original model consisting of all directional attention paths and $n_h\shortequal 3$ is used. The results are summarized in Table \ref{tab:ablation}.
The quality of estimated texture maps ($G_{\cT}$), evaluated by rendered images, improves when image-to-texture flows are added to the independent model, and improves further when texture-to-image flows are incorporated. For the image generator ($G_{\cI}$), neither the $\cI\shortrarrow\cT$ nor $\cT\shortrarrow\cI$ model shows a consistent improvement, however, the two types of flow altogether improve the overall generation quality. Figure \ref{fig:ablation} shows some rendered images comparing ablated models. In Figure \ref{fig:ablation}, the \textit{Ind.} and the $\cT\shortrarrow\cI$ models cannot generate distinguishable black sleeves in the first example while $\cI\shortrarrow\cT$ model and the full model generate distinguishable black sleeves. The \textit{Ind.} and $\cT\shortrarrow\cI$ models often fail at generating accurate clothes color while the full model succeed. The $\cI\shortrarrow\cT$ model generates comparable texture map to the full model, however, the it often generates some artifacts on texture map and background.

\begin{figure}[t]
    \centering
    \includegraphics[width=\columnwidth]{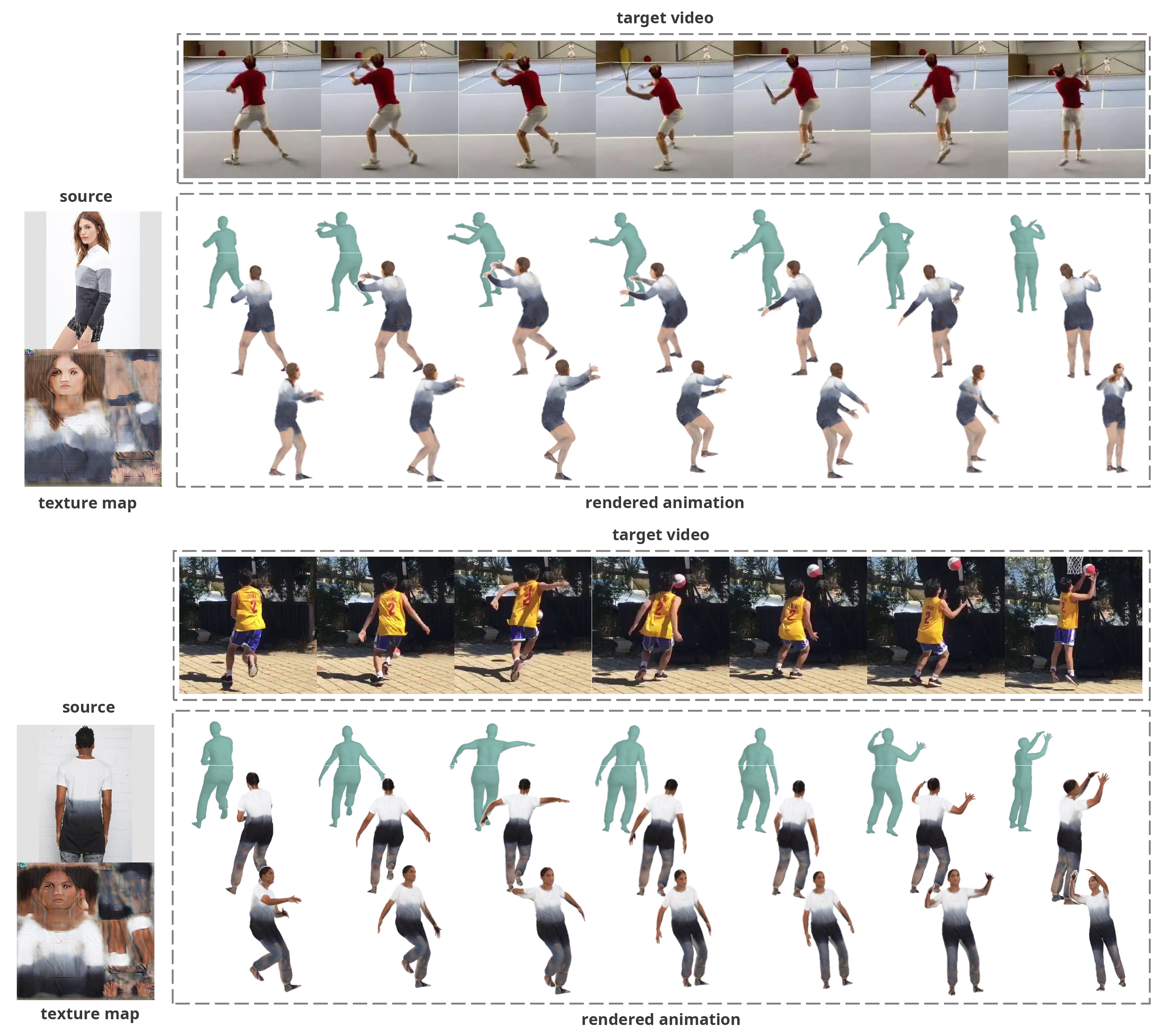}
    \caption{Examples of 3D animation rendering using our estimated texture map viewed in two different viewpoints.}
    \label{fig:application}
\end{figure}

\subsection{Application}

To verify the usability of our method for 3D model rendering, we generate 3D animation clips using texture maps generated by our method. We first reconstruct a sequence of 3D human shape in the SMPL format using the off-the-shelf 3D video reconstruction model \citep{kocabas2020vibe} and the off-the-shelf clothing model \citep{ma2020learning}. Then we apply a texture map generated by ours to the 3D shape sequence to obtain a colored animation clip. Figure \ref{fig:application} shows some examples. The generated 3D animations are viewed in two different viewpoints. The result shows that a person in a source image acts as the target target video, preserving clothes patterns all around. The generated texture map works effectively for 3D model rendering, generating consistent images for any pose and viewpoint.

\section{Conclusion}

We propose dual-domain generative models for a complete texture map estimation by providing multi-view features using a novel hallucination generation scheme. Our model utilizes a local attention module over the domains to convey multi-view features to the texture map and texture features to pose transferred images. Experimental results show that the estimated texture map has decent quality for rendering colorful 3D human models, which is applicable to generate a free-view point 3D animation.

\section*{Acknowledgment}

This work was supported by Institute of Information \& Communications Technology Planning \& Evaluation (IITP) grant funded by the Korea government (MSIT) (No. 2019-0-01190, [SW Star Lab] Robot Learning: Efficient, Safe, and Socially-Acceptable Machine Learning).

\bibliography{egbib}

\begin{thebibliography}{39}
\providecommand{\natexlab}[1]{#1}

\bibitem[{Alldieck et~al.(2019)Alldieck, Magnor, Bhatnagar, Theobalt, and
  Pons-Moll}]{alldieck2019learning}
Alldieck, T.; Magnor, M.; Bhatnagar, B.~L.; Theobalt, C.; and Pons-Moll, G.
  2019.
\newblock Learning to Reconstruct People in Clothing From a Single {RGB}
  Camera.
\newblock In \emph{IEEE Conf. Comput. Vis. Pattern Recog.}

\bibitem[{Bogo et~al.(2016)Bogo, Kanazawa, Lassner, Gehler, Romero, and
  Black}]{bogo2016keep}
Bogo, F.; Kanazawa, A.; Lassner, C.; Gehler, P.; Romero, J.; and Black, M.~J.
  2016.
\newblock Keep It {SMPL:} Automatic Estimation of 3D Human Pose and Shape from
  a Single Image.
\newblock In \emph{Eur. Conf. Comput. Vis.} Springer.

\bibitem[{Cao et~al.(2019)Cao, Hidalgo, Simon, Wei, and
  Sheikh}]{cao2019openpose}
Cao, Z.; Hidalgo, G.; Simon, T.; Wei, S.-E.; and Sheikh, Y. 2019.
\newblock OpenPose: Realtime Multi-Person 2D Pose Estimation Using Part
  Affinity Fields.
\newblock \emph{IEEE Trans. Pattern Anal. Mach. Intell.}, 43(1): 172--186.

\bibitem[{Catmull(1974)}]{catmull1974subdivision}
Catmull, E. 1974.
\newblock \emph{A Subdivision Algorithm for Computer Display of Curved
  Surfaces}.
\newblock Ph{D} dissertation, Utah Univ Salt Lake City School of Computing.

\bibitem[{Catmull and Smith(1980)}]{catmull19803}
Catmull, E.; and Smith, A.~R. 1980.
\newblock 3-D Transformations of Images in Scanline Order.
\newblock In \emph{SIGGRAPH}, 279--285. ACM.

\bibitem[{Choutas et~al.(2020)Choutas, Pavlakos, Bolkart, Tzionas, and
  Black}]{choutas2020monocular}
Choutas, V.; Pavlakos, G.; Bolkart, T.; Tzionas, D.; and Black, M.~J. 2020.
\newblock Monocular Expressive Body Regression Through Body-Driven Attention.
\newblock In \emph{Eur. Conf. Comput. Vis.} Springer.

\bibitem[{Gabeur et~al.(2019)Gabeur, Franco, Martin, Schmid, and
  Rogez}]{gabeur2019moulding}
Gabeur, V.; Franco, J.-S.; Martin, X.; Schmid, C.; and Rogez, G. 2019.
\newblock Moulding Humans: Non-Parametric 3D Human Shape Estimation From Single
  Images.
\newblock In \emph{Int. Conf. Comput. Vis.}

\bibitem[{Grigorev et~al.(2019)Grigorev, Sevastopolsky, Vakhitov, and
  Lempitsky}]{grigorev2018coordinate}
Grigorev, A.; Sevastopolsky, A.; Vakhitov, A.; and Lempitsky, V. 2019.
\newblock Coordinate-Based Texture Inpainting for Pose-Guided Human Image
  Generation.

\bibitem[{G{\"u}ler, Neverova, and Kokkinos(2018)}]{guler2018densepose}
G{\"u}ler, R.~A.; Neverova, N.; and Kokkinos, I. 2018.
\newblock DensePose: Dense Human Pose Estimation in the Wild.
\newblock In \emph{IEEE Conf. Comput. Vis. Pattern Recog.}

\bibitem[{Heusel et~al.(2017)Heusel, Ramsauer, Unterthiner, Nessler, and
  Hochreiter}]{heusel2017gans}
Heusel, M.; Ramsauer, H.; Unterthiner, T.; Nessler, B.; and Hochreiter, S.
  2017.
\newblock Gans trained by a two time-scale update rule converge to a local nash
  equilibrium.
\newblock In \emph{Adv. Neural Inform. Process. Syst.}

\bibitem[{Jian et~al.(2019)Jian, Yunshan, Yachun, Chi, and
  Yichen}]{wang2019reidsupervised}
Jian, W.; Yunshan, Z.; Yachun, L.; Chi, Z.; and Yichen, W. 2019.
\newblock Re-Identification Supervised Texture Generation.
\newblock \emph{IEEE Conf. Comput. Vis. Pattern Recog.}

\bibitem[{Johnson, Alahi, and Fei-Fei(2016)}]{johnson2016perceptual}
Johnson, J.; Alahi, A.; and Fei-Fei, L. 2016.
\newblock Perceptual Losses for Real-Time Style Transfer and Super-Resolution.
\newblock In \emph{Eur. Conf. Comput. Vis.} Springer.

\bibitem[{Kanazawa et~al.(2018)Kanazawa, Black, Jacobs, and
  Malik}]{kanazawa2018end}
Kanazawa, A.; Black, M.~J.; Jacobs, D.~W.; and Malik, J. 2018.
\newblock End-to-End Recovery of Human Shape and Pose.
\newblock In \emph{IEEE Conf. Comput. Vis. Pattern Recog.}

\bibitem[{Kocabas, Athanasiou, and Black(2020)}]{kocabas2020vibe}
Kocabas, M.; Athanasiou, N.; and Black, M.~J. 2020.
\newblock {VIBE:} Video Inference for Human Body Pose and Shape Estimation.
\newblock In \emph{IEEE Conf. Comput. Vis. Pattern Recog.}

\bibitem[{Kolotouros et~al.(2019)Kolotouros, Pavlakos, Black, and
  Daniilidis}]{kolotouros2019spin}
Kolotouros, N.; Pavlakos, G.; Black, M.~J.; and Daniilidis, K. 2019.
\newblock Learning to Reconstruct 3D Human Pose and Shape via Model-Fitting in
  the Loop.
\newblock In \emph{Int. Conf. Comput. Vis.}

\bibitem[{Lassner et~al.(2017)Lassner, Romero, Kiefel, Bogo, Black, and
  Gehler}]{lassner2017unite}
Lassner, C.; Romero, J.; Kiefel, M.; Bogo, F.; Black, M.~J.; and Gehler, P.~V.
  2017.
\newblock Unite the People: Closing the Loop Between 3D and 2D Human
  Representations.
\newblock In \emph{IEEE Conf. Comput. Vis. Pattern Recog.}

\bibitem[{Lazova, Insafutdinov, and Pons-Moll(2019)}]{lazova2019360}
Lazova, V.; Insafutdinov, E.; and Pons-Moll, G. 2019.
\newblock 360-Degree Textures of People in Clothing from a Single Image.
\newblock In \emph{Int. Conf. 3D Vision (3DV)}. IEEE.

\bibitem[{Liu et~al.(2019)Liu, Piao, Min, Luo, Ma, and Gao}]{liu2019liquid}
Liu, W.; Piao, Z.; Min, J.; Luo, W.; Ma, L.; and Gao, S. 2019.
\newblock Liquid Warping {GAN:} {A} Unified Framework for Human Motion
  Imitation, Appearance Transfer and Novel View Synthesis.
\newblock In \emph{Int. Conf. Comput. Vis.}

\bibitem[{Liu et~al.(2016)Liu, Luo, Qiu, Wang, and Tang}]{liu2016deepfashion}
Liu, Z.; Luo, P.; Qiu, S.; Wang, X.; and Tang, X. 2016.
\newblock DeepFashion: Powering Robust Clothes Recognition and Retrieval with
  Rich Annotations.
\newblock In \emph{IEEE Conf. Comput. Vis. Pattern Recog.}

\bibitem[{Loper et~al.(2015)Loper, Mahmood, Romero, Pons-Moll, and
  Black}]{loper2015smpl}
Loper, M.; Mahmood, N.; Romero, J.; Pons-Moll, G.; and Black, M.~J. 2015.
\newblock {SMPL}: A Skinned Multi-Person Linear Model.
\newblock \emph{ACM Trans. Graph.}, 34(6): 248:1--248:16.

\bibitem[{Ma et~al.(2017)Ma, Jia, Sun, Schiele, Tuytelaars, and
  Van~Gool}]{ma2017pose}
Ma, L.; Jia, X.; Sun, Q.; Schiele, B.; Tuytelaars, T.; and Van~Gool, L. 2017.
\newblock Pose Guided Person Image Generation.
\newblock In \emph{Adv. Neural Inform. Process. Syst.}

\bibitem[{Ma et~al.(2020)Ma, Yang, Ranjan, Pujades, Pons-Moll, Tang, and
  Black}]{ma2020learning}
Ma, Q.; Yang, J.; Ranjan, A.; Pujades, S.; Pons-Moll, G.; Tang, S.; and Black,
  M.~J. 2020.
\newblock Learning to Dress 3D People in Generative Clothing.
\newblock In \emph{IEEE Conf. Comput. Vis. Pattern Recog.}

\bibitem[{Natsume et~al.(2019)Natsume, Saito, Huang, Chen, Ma, Li, and
  Morishima}]{natsume2019siclope}
Natsume, R.; Saito, S.; Huang, Z.; Chen, W.; Ma, C.; Li, H.; and Morishima, S.
  2019.
\newblock SiCloPe: Silhouette-Based Clothed People.
\newblock In \emph{IEEE Conf. Comput. Vis. Pattern Recog.}

\bibitem[{Neverova, Alp~Guler, and Kokkinos(2018)}]{neverova2018dense}
Neverova, N.; Alp~Guler, R.; and Kokkinos, I. 2018.
\newblock Dense Pose Transfer.
\newblock In \emph{Eur. Conf. Comput. Vis.}

\bibitem[{Pavlakos et~al.(2018)Pavlakos, Zhu, Zhou, and
  Daniilidis}]{pavlakos2018learning}
Pavlakos, G.; Zhu, L.; Zhou, X.; and Daniilidis, K. 2018.
\newblock Learning to Estimate 3D Human Pose and Shape From a Single Color
  Image.
\newblock In \emph{IEEE Conf. Comput. Vis. Pattern Recog.}

\bibitem[{Ren et~al.(2020)Ren, Yu, Chen, Li, and Li}]{ren2020deep}
Ren, Y.; Yu, X.; Chen, J.; Li, T.~H.; and Li, G. 2020.
\newblock Deep Image Spatial Transformation for Person Image Generation.
\newblock In \emph{IEEE Conf. Comput. Vis. Pattern Recog.}

\bibitem[{Saito et~al.(2019)Saito, Huang, Natsume, Morishima, Kanazawa, and
  Li}]{saito2019pifu}
Saito, S.; Huang, Z.; Natsume, R.; Morishima, S.; Kanazawa, A.; and Li, H.
  2019.
\newblock PIFu: Pixel-Aligned Implicit Function for High-Resolution Clothed
  Human Digitization.
\newblock In \emph{Int. Conf. Comput. Vis.}

\bibitem[{Saito et~al.(2020)Saito, Simon, Saragih, and Joo}]{saito2020pifuhd}
Saito, S.; Simon, T.; Saragih, J.; and Joo, H. 2020.
\newblock PIFuHD: Multi-Level Pixel-Aligned Implicit Function for
  High-Resolution 3D Human Digitization.
\newblock In \emph{IEEE Conf. Comput. Vis. Pattern Recog.}

\bibitem[{Siarohin et~al.(2018)Siarohin, Sangineto, Lathuilière, and
  Sebe}]{siarohin2018deformable}
Siarohin, A.; Sangineto, E.; Lathuilière, S.; and Sebe, N. 2018.
\newblock Deformable GANs for Pose-Based Human Image Generation.
\newblock In \emph{IEEE Conf. Comput. Vis. Pattern Recog.}

\bibitem[{Thorm{\"a}hlen and Seidel(2008)}]{thormahlen20083d}
Thorm{\"a}hlen, T.; and Seidel, H.-P. 2008.
\newblock 3D-Modeling by Ortho-Image Generation from Image Sequences.
\newblock \emph{ACM Trans. Graph.}, 27(3): 1--5.

\bibitem[{Varol et~al.(2018)Varol, Ceylan, Russell, Yang, Yumer, Laptev, and
  Schmid}]{varol2018bodynet}
Varol, G.; Ceylan, D.; Russell, B.; Yang, J.; Yumer, E.; Laptev, I.; and
  Schmid, C. 2018.
\newblock BodyNet: Volumetric Inference of 3D Human Body Shapes.
\newblock In \emph{Eur. Conf. Comput. Vis.} Springer.

\bibitem[{Vaswani et~al.(2017)Vaswani, Shazeer, Parmar, Uszkoreit, Jones,
  Gomez, Kaiser, and Polosukhin}]{vaswani2017attention}
Vaswani, A.; Shazeer, N.; Parmar, N.; Uszkoreit, J.; Jones, L.; Gomez, A.~N.;
  Kaiser, L.; and Polosukhin, I. 2017.
\newblock Attention is All you Need.
\newblock In \emph{Adv. Neural Inform. Process. Syst.}

\bibitem[{Wang et~al.(2001)Wang, Kang, Szeliski, and Shum}]{wang2001optimal}
Wang, L.; Kang, S.~B.; Szeliski, R.; and Shum, H.-Y. 2001.
\newblock Optimal Texture Map Reconstruction from Multiple Views.
\newblock In \emph{IEEE Conf. Comput. Vis. Pattern Recog.}

\bibitem[{Wang et~al.(2004)Wang, Bovik, Sheikh, and Simoncelli}]{wang2004image}
Wang, Z.; Bovik, A.~C.; Sheikh, H.~R.; and Simoncelli, E.~P. 2004.
\newblock Image Quality Assessment: From Error Visibility to Structural
  Similarity.
\newblock \emph{IEEE Trans. Image Process.}, 13(4): 600--612.

\bibitem[{Weng, Curless, and Kemelmacher-Shlizerman(2019)}]{weng2019photo}
Weng, C.-Y.; Curless, B.; and Kemelmacher-Shlizerman, I. 2019.
\newblock Photo Wake-Up: 3D Character Animation From a Single Photo.
\newblock In \emph{IEEE Conf. Comput. Vis. Pattern Recog.}

\bibitem[{Zablotskaia et~al.(2019)Zablotskaia, Siarohin, Zhao, and
  Sigal}]{zablotskaia2019dwnet}
Zablotskaia, P.; Siarohin, A.; Zhao, B.; and Sigal, L. 2019.
\newblock DwNet: Dense warp-based network for pose-guided human video
  generation.
\newblock In \emph{Brit. Mach. Vis. Conf.}

\bibitem[{Zhang et~al.(2018)Zhang, Isola, Efros, Shechtman, and
  Wang}]{zhang2018unreasonable}
Zhang, R.; Isola, P.; Efros, A.~A.; Shechtman, E.; and Wang, O. 2018.
\newblock The Unreasonable Effectiveness of Deep Features as a Perceptual
  Metric.
\newblock In \emph{IEEE Conf. Comput. Vis. Pattern Recog.}

\bibitem[{Zhao et~al.(2020)Zhao, Liao, Zhang, and Shao}]{zhao2020human}
Zhao, F.; Liao, S.; Zhang, K.; and Shao, L. 2020.
\newblock Human Parsing Based Texture Transfer from Single Image to 3D Human
  via Cross-View Consistency.
\newblock In \emph{Adv. Neural Inform. Process. Syst.}

\bibitem[{Zhi et~al.(2020)Zhi, Lassner, Tung, Stoll, Narasimhan, and
  Vo}]{zhi2020texmesh}
Zhi, T.; Lassner, C.; Tung, T.; Stoll, C.; Narasimhan, S.~G.; and Vo, M. 2020.
\newblock TexMesh: Reconstructing Detailed Human Texture and Geometry from
  RGB-D Video.
\newblock In \emph{Eur. Conf. Comput. Vis.} Springer.

\end{thebibliography}

\end{document}